# Person Re-Identification System at Semantic Level based on Pedestrian Attributes Ontology

Ngoc Q. Ly[1], Hieu N. M. Cao[2]
Department of Computer Vision and Cognitive Cybernetics
VNUHCM-University of Science
Ho Chi Minh City
Vietnam

Thi T. Nguyen[3]
Computer Vision and Cognitive Cybernetics
VNUHCM-University of Science
Ho Chi Minh City
Vietnam

*Abstract*—**Person Re-Identification (Re-ID) is a very important task in video surveillance systems such as tracking people, finding people in public places, or analysing customer behavior in supermarkets. Although there have been many works to solve this problem, there are still remaining challenges such as large-scale datasets, imbalanced data, viewpoint, fine-grained data (attributes), the Local Features are not employed at semantic level in online stage of Re-ID task, furthermore, the imbalanced data problem of attributes are not taken into consideration. This paper has proposed a Unified Re-ID system consisted of three main modules such as Pedestrian Attribute Ontology (PAO), Local Multi-task DCNN (Local MDCNN), Imbalance Data Solver (IDS). The new main point of our Re-ID system is the power of mutual support of PAO, Local MDCNN and IDS to exploit the inner-group correlations of attributes and pre-filter the mismatch candidates from Gallery set based on semantic information as Fashion Attributes and Facial Attributes, to solve the imbalanced data of attributes without adjusting network architecture and data augmentation. We experimented on the well-known Market1501 dataset. The experimental results have shown the effectiveness of our Re-ID system and it could achieve the higher performance on Market1501 dataset in comparison to some state-of-the-art Re-ID methods.**

## I. INTRODUCTION

Re-ID is the problem of recognising and associating a person at different physical locations over time after the person had been previously observed visually elsewhere. Solving the Re-ID problem has gained a rapid increase in attention in both academic research communities and industrial laboratories in recent years. It has many applications, such as tracking people across cameras, images retrieval, or customer behavior analysis [1]. Due to using appearance features from input images, this problem suffers from the common challenges in visual recognition: illumination, pose variation, occlusion, intra-class and inter-class variations. Early studies aim to make full use of hand-crafted visual features [2-8] or metric learning [2,4,5, 9-11] to build a best descriptor for each person. These methods can solve one or some of the above challenges, but are very computational expensive and do not reach high results of accuracy. In recent years, with the growth of convolutional neural networks (CNNs) and careful-annotated benchmarks, CNN-based models which learn deep features from data outperform hand-crafted methods by a large margin and achieve remarkable accuracy [12]. To obtain more discriminative features to deal with the inter-class challenge, some works try to extract local features from local regions in different ways, such as pose normalization [13-15], part-based learning [16-19], or attention mechanism [20-23]. Although these great works gain very high performance in accuracy and mAP, they are still only employed deep features, which do not contain semantic information and cannot be explained by human.

Pedestrian Attribute Recognition is a task that predicts a number of predefined attributes describing a pedestrian. Similar to Re-ID, this task takes bounding boxes of pedestrian captured by cameras as inputs. Attributes are semantic information. They are extracted based on attribute learning model. They could be robust to challenges such as pose, lighting, camera characteristics. According to Re-ID problem, facial attributes and cloth attributes are considered. Combining a set of large enough attributes can help improve the discrimination of Re-ID features. Furthermore, unlike low-level visual features or high-level deep features, attributes are easy to understand for human [24]. Attributes can also be expanded to a range of other applications, such as clothes retrieval, face retrieval. Most existing Re-ID studies use global features to predict all attributes [24-27]. However, most attributes appear in local positions, so global features are insufficient to recognize them. Some works notice this drawback and improve by divide global features into local parts [28,29], but they still consider attributes as an auxiliary branch to enrich deep features.

In this work, we proposed two simple CNN-based models, one for extracting deep global features, and the other for predicting pedestrian attributes. In the learning stage of the attribute recognition model (ARM), we split feature maps at a specific mid-level layer into multiple branches, with respect to human's body parts. Each branch use a local feature map, which is horizontally split from global feature map, to predict a group of corresponding relevant attributes. The attributes groups are applied from a predefined PAO, which can help leverage the intra-class correlation of attributes into the learning process. Besides, we take into consideration the imbalance of attributes and handle it by employing the Matthews correlation coefficient (MCC). In the inference stage, different from previous methods, for each query image, we firstly use attributes prediction to filter out mismatch candidates, and then the remaining ones will be used to find out best matching by deep global features.

The main contributions of this paper are as follows: 1) We propose the Pedestrian Attribute Ontology to conduct Pedestrian Attribute Learning Process and Re-ID process; 2) We propose the Pedestrian Attribute Learning Model based on Local Multi-task learning; 3) We propose integrating Imbalanced Data Solver based on MCC to Re-ID system; 4) We propose new Re-ID method based on Deep Global Features and Pedestrian Semantic Information.

## II. RELATED WORKS

### A. Hand-Crafted-Features-based Re-ID

Traditional approaches mainly focus on designing discriminative visual hand-crafted features. Colors and texture are usually employed. In [2], RGB and HSV color vectors are extracted from input images and then fed into a Maximum Likelihood model to learn the image similarities. In another approach [3], Gabor filters are used to extract texture features. Covariances of these features are also employed by the Region Covariance Descriptor in [6,7]. In [8], a robust features named Local Maximal Occurrence (LOMO) are proposed. LOMO is obtained by sliding a window on input images and taking the maximum values of different features from all patches under the window. Apart from proposing discriminative features, other works [5, 9-11] try to design an effective metric to learn the similarity and difference between images.

### B. CNN-based Re-ID

CNNs have been widely employed in person re-identification due to their great performance in many different computer vision tasks [30–32]. Earlier works use global features extracted from a CNN to train a siamese network [33–35]. For example, [33] proposed a Deep Ranking model aiming to maximize the rank of Euclide distance of same identity's feature vectors. Author in [34] employ Recurrent Neural Network to make use of motion information for more discriminative person description. Author in [35] proposed a Pyramid Person Matching Network to learn the correspondence of misalignment components in image pairs. Attention mechanisms are applied in many models [20-23] to focus on salient parts to extract more useful and discriminative information. Recent studies start to consider part-level features as complementary information for their models due to its fine-grained information. Early part-based approaches apply predefined rigid grids on input images as local parts [36,37]. This way of partition is insufficient because person detection boxes are not always correct. In a very detail partition way, [38] train a semantic parsing model to localize pixel-level body parts. A weighted sum layer is then used to fuse global and local features for identity classification. Extra pose estimators [13,14] or spatial constraints [15] are also utilized to normalize deformable pedestrian parts to obtain more robust features. In another way to learn local features [16-18], global features are horizontal pooled and then separated into vectors, each of them is considered as containing information of a correspond body part. Author in [17] do the same methods but splitted global features into equal stripes in multiple granularities. Each stripe is then adopted separately to an identity classification. Many of the above methods achieve remarkable performance in Re-ID task. However, none of them consider semantic information, such as attributes, but only try to make robust deep global/local features.

### C. Attribute for Re-ID

Attributes are signatures of concepts. It is difficult to recognize concepts, but recognizing signatures is much easier. Attributes help re-identify pedestrians from coarse to fine, and help to understand images in a detail level. Therefore, attributes can help increase the discrimination between pedestrians. There are many works investigating attributes as auxiliary information to Re-ID. In [24,25], a DCNN classifier is first trained on an independent attributes annotated dataset, then the attributes predicted by that model is used to train and fine-tune another person re-id model. It can be noticed that this kind of methods would push the error of former attribute model to the Re-ID model, especially the attribute model still do not consider the imbalance problem. In [26,27], an end-to-end Multi-task DCNN model is proposed to do both attribute recognition and Re-ID tasks simultaneously. In these works, each attribute probability is predicted by forwarding a same global feature vector to a corresponding linear layer. This vector is also used to retrieve nearest neighbors in inference stage of Re-ID task. In fact, many attributes just appear in small regions on human body, so using a unique global feature vector to learn all attributes is inefficient. Recognizing this drawback, [28,29] proposed part-based CNN models, in which a global feature map from a middle layer is horizontally split into 4 disjoint equal local feature maps, each one is then forwarded to some other convolutional layers followed by a last linear layer to predict probabilities of a group of attributes. The improvement in this method is that it use multiple local features to predict groups of suitable attributes. However, some attributes are distributed over more than one part, so it is confused to choose output of which part to evaluate attributes recognition. Therefore, in inference stage of Re-ID task, the authors do not use attributes predictions anymore, but only enrich features by concatenating all deep local and global features. Furthermore, none of the above methods handle the imbalance data problem of attributes, which is an inherent problem in many classification tasks.

Therefore, in this paper, followed the methods in [28], which is to build a DCNN model that split middle global feature map into multiple local parts. But instead of distributing each attribute over multiple parts and concatenating local and global deep features, we proposed novel methods for improvement: 1) We build a Pedestrian Attribute Ontology (PAO) for better attributes learning, and also for easily expanding in the future; 2) Based on PAO, we build a Local Multi-task DCNN model (Local MDCNN) to exploit inner group and inter group correlations between attributes; 3) We incorporate an Imbalanced Data Solver (IDS) to our Pedestrian Attribute Recognition module; and 4) we build a novel Person Re-identification system flexibly combining global deep features and pedestrian semantic information (facial and cloth attributes).

## III. METHOD

Briefly, our contribution is a novel Person Re-identification system based on Deep Global Features and Pedestrian Attributes. In this section, we focus on the main points of our system. In the off-line stage, we build a PAO and then a Local MDCNN to learn pedestrian attributes. Besides, we use transfer learning to train a siamese network to learn person global deep features. In the on-line stage, for each query image, we firstly use pedestrian semantic information, so-called attribute, to pre-filter candidate images, and then use global deep features to find nearest neighbor in the remaining ones, or vice versa.

Our Re-ID plays a very important role in a multi-camera tracking person system. In each viewing range of camera, the tracking task can be performed by the conventional methods, but when the person moves from view range of camera (i) to camera (i+1), Re-ID is very useful to identify the monitored person is being lost track.

Our system is organized into two phases: *1) Offline Phase:* This phase is designed to build the PAO to support the Deep Global Features Learning model (DGFL model) and the Pedestrian Attributes Learning model (PAL model, aka the Local MDCNN model). In order to improve the performance of learning models, we take into account the imbalanced data problem, and to prepare features for gallery set for matching process in the online phase. The PAO is manually built based on domain knowledge in the field of fashion attributes and facial attributes. It is a hierarchical semantic tree. Its purpose is to exploit the correlations between attributes for not only better learning but also easier expanding in the future. The DGFL model is designed and trained to extract the deep global features of the input image of each person. The PAL model is designed and trained to extract the predefined attributes of each person. The deep features and the semantic information such as attributes are then mutually combined in the online phase. The imbalanced data problem also is solved in this phase by the IDS. Its purpose is to find the best thresholds for each attribute prediction in the training set. In the online phase, the best chosen thresholds are used to convert continuous predicted outputs of PAL model to corresponding binary values to use for the query process. *2) Online Phase:* This phase is organized to run the query process including deep features extraction, attributes information extraction and retrieval. After getting deep features from DGFL model and attributes information from PAL model, for each query image, we firstly use attributes to filter out candidate images with different attributes of the query image, and then use global deep features to find nearest neighbor in the remain ones, or do the steps vice versa.

### A. Pedestrian Attribute Ontology (PAO)

Our PAO is inspired by our Face Attribute Ontology (FaAO) and our Fashion Attribute Ontology (FasAO) [39]. PAO helps us to exploit inner group and inter group correlations between attributes, it is then very useful for training the Local MDCNN. PAO also helps the developer to easily update new attributes in the future. To do this, we firstly manually classify attributes into groups based on their corresponding correlations. These groups are combined into a semantic hierarchical tree, also known as attributes ontology. To bring the PAO into deep model, we base on an observation that, most of attributes can be seen at a local region instead of a whole human body. For example, "is wearing hat" can be predicted by learning from "head" region. Therefore, our Local MDCNN learns attributes from local features instead of global ones. Follow [39], we employ a clothing attribute dataset named DeepFashion [40] to build a general FasAO, and our prior knowledge to build a general FaAO. In experiment, we re-build specific PAO on another dataset that has both Re-ID and attribute label.

In [41], Gruber stated that ontology is a formal, explicit specification of shared concepts. An ontology is formed by four principle components: individuals, classes, concepts and relations between them. The class components can has multiple layers. In our case, we formulate person attribute ontology by five components:

- Person (individuals): a layer representing people objects.

- Regions (classes): a layer representing human's body regions, consisted of five parts: head, upper body, lower body, whole body (upper and lower) and foot.

- Categories (classes): a layer representing types of corresponding clothes in each region.

- Attributes (concepts): clothing attributes with respect to each category and facial attributes.

- Relations: consisted of 3 relations: part of (between regions and individuals), has a (between regions - categories and categories - attributes), is a (between attributes and their values)

The semantic hierarchical tree consisted of three main levels: Regions, Categories and Attributes. Fig. 1 shows our PAO. In Fig. 1, human body firstly is split into five regions. In each region, there are multiple categories of clothing items. And for each item, there are relevant attributes depending on its kind. Basically, it is not too difficult to know which items should be put into which body regions. Here we take some examples from DeepFashion dataset and visualize in Fig. 2 to show some popular kinds of clothing items. The PAO shows two properties of attributes which are inner group correlation and inter group correlation. These two properties help us in the step of designing deep model that are:

- Firstly, when training a deep attribute recognition model, global features are usually be used to predict all attributes. But, in real life, people just need to see a local region to find out attributes related to this region. For example, we can know if a man is wearing hat or not by looking at his head, and do not need to look at the other regions. This is the inter group correlation between attributes. Ontology help us to see which attributes should not go together and therefore should not be predicted in same local features.

- Secondly, there are many attributes having co-appearance relation to each other. For example, a person having beard is usually a man, so the two attributes Is Male and Having beard usually being zeros or ones together. This is the inner group correlation. Ontology help us to group these attributes into same classes, and therefore deep model should predict them in same local features

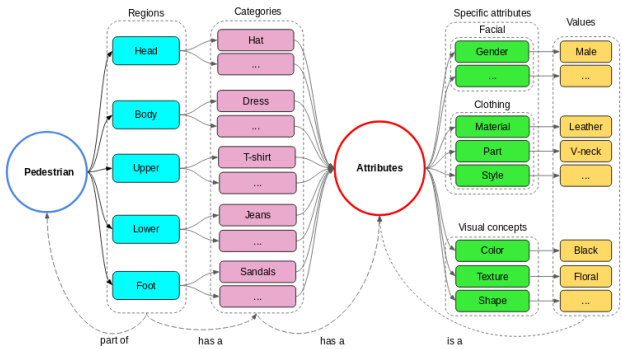

Fig. 1. Our Pedestrian Attribute Ontology.

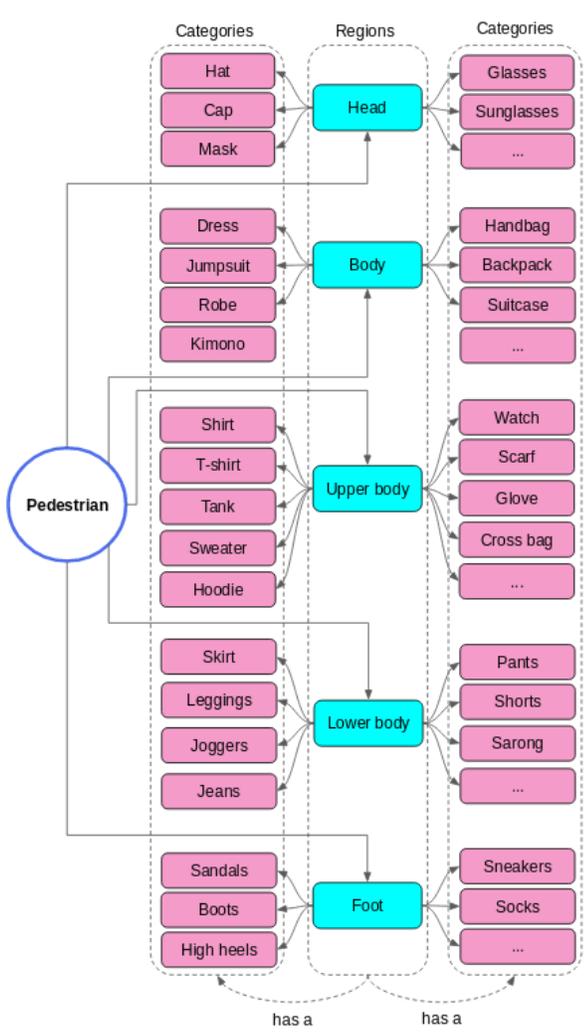

Fig. 2. Some kinds of Clothing Items Extracted from the PAO.

Clothing attributes are very numerous and variety. Follow Ly et al. we choose six types of clothing attributes to demonstrate our ontology, and divide them into two groups: i) general attributes which are attributes that most of items would have, include: color, texture, shape; and ii) specific attributes which are attributes that only exist on some items. Besides, we also show some facial attributes, which can be recognized from the head position. Some examples for clothing attributes and facial attributes are showed in Fig. 3 and Fig. 4, respectively.

In summary, Pedestrian Attribute Ontology is a hierarchical semantic tree, in which attributes are classified into groups. It not only can improve learning process of deep model (in comparison to the models without it), but also easily update more attributes if it is necessary in the future. In the next sub-section, we base on this ontology to design an effective deep model for attribute recognition task.

*1) Models:* Our system bases on two models: a Person Deep Global Features Learning model and a Pedestrian Attribute Learning model. The former one is trained to extract deep global features vectors and the latter one is trained to extract attributes vectors. Since deep features achieved high performances on many tasks in computer vision, it should not be ignored in our system. However, deep features do not contain semantic information. With only one input image, we cannot understand what do deep features mean, but with attributes features, we can know which attributes exist on the persons in the input images. Therefore, both of the above features can mutually support to get high performance in our system.

*a) Person Deep Global Features Learning model:* Since we are trying to proof by experiments that attributes prediction can help improve Re-ID results, so we just build a simple person deep global features learning model instead of using complex architecture like other great works. Concretely, we transfer 50-layers Residual Network [32] which was pretrained on the famous image classification dataset ImageNet. In our architecture, we remove the last 1000-units linear layer and append a 1x1 convolutional layer to reduce dimension from 2048 down to 256. This last 256-D vector is the feature vector of the input bounding box, which is then used for matching in the inference stage. Fig. 5 show our Re-ID model architecture in the offline phase.

A couple of images are of the same person when their corresponding deep features have high similarity. To train the model to achieve this goal, in training stage, we use Triplet Loss [42] as our loss function. This function was used by many Re-ID models in particular and image retrieval models in general. Its goal is to learn the similarity between same ID inputs and the divergence between different ID inputs. Equation (1) shows the formula of this function. Whenever the training process is finished, Euclidean dissimilarity distance between feature vectors of images of same person ($f^a$ and $f^p$) should less than those of different person ($f^a$ and $f^n$) by a margin m.

$$\text{TripletLoss} = \max(0, \parallel f^a - f^p \parallel_2^2 - \parallel f^a - f^n \parallel_2^2 + m \quad (1)$$

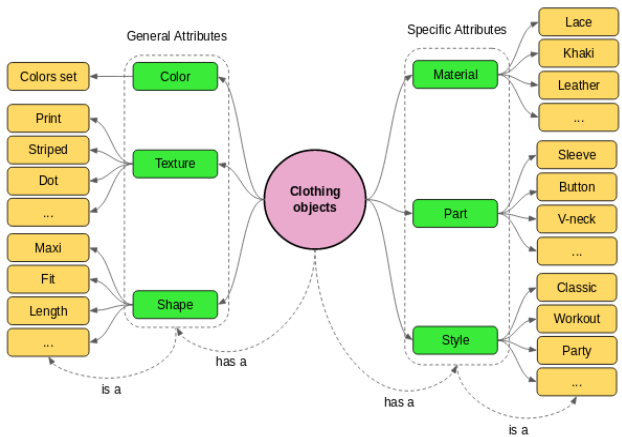

Fig. 3. Some Clothing Attributes with their Values, Extracted from the PAO.

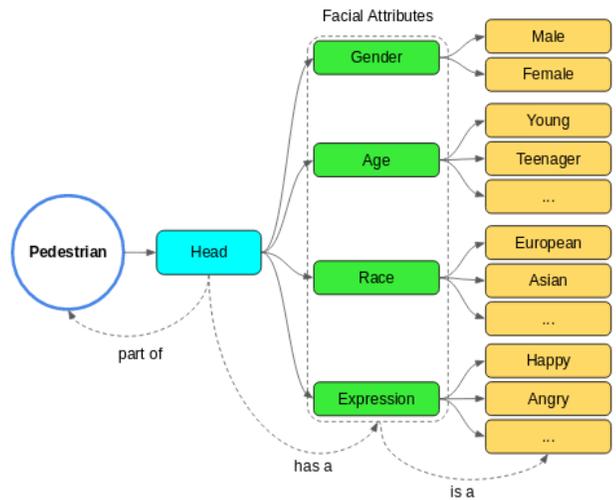

Fig. 4. Some Facial Attributes with their Values, Extracted from the PAO.

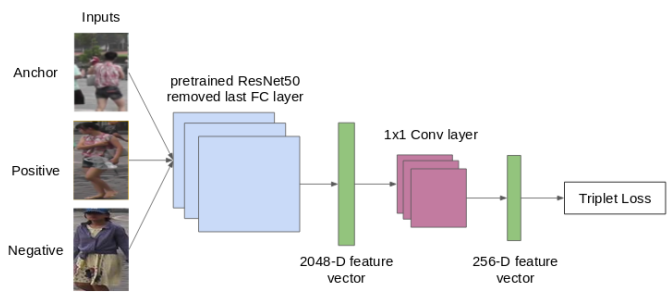

Fig. 5. Person Deep Global Features Learning Model. The Inputs of each Timestep Are a Triple of Images, Including Anchor, Positive and Negative one. The Model Parameters are Shared between them.

In the inference stage, with an input image, its corresponding deep features vectors are extracted and then compared to other pre-extracted vectors of gallery images. The nearest gallery one, i.e. the one has smallest Euclidean dissimilarity distance, would be chosen to match with input image as the same person. With the above strategy, the more similar the two individuals are, the smaller Euclidean dissimilarity distance between the corresponding deep features vectors has. However, this leads to another drawback of deep features: mismatched results would rise in cases of different persons having same appearance (such as same cloths, same pose). In these cases, we need more detail information to distinguish them instead of global deep features only. And pedestrian attributes are semantic information at fine-grained level. Therefore, the attributes are used in our system to filter out false positive candidate images in those cases to get better performance for Re-ID system.

*b) Person Attribute Learning model:* Pedestrian Attribute Learning model is designed based on Pedestrian Attribute Ontology. From the ontology, to make sure that the inner group and inter group correlation can be leveraged into deep model, we build PAL model with three levels: regions, categories and attributes. Firstly, a middle layer global feature map is split into four equal parts, which means each one occupies 25% ratio height of person body. From top to bottom, the parts respectively are head, upper body, lower body, foot region. The body region is the merger of upper body and lower body. Secondly, each region split features are learnt in a corresponding local sub-network to extract information relevant to that local part. And finally, local features of each part are fed into multiple smaller branches, which then predict the attributes in the PAO.

In briefly, our PAL model has three parts: i) a global part that learns common features; multiple attribute learning sub-networks, consist of: ii) local parts that learn local features, and iii) attribute parts that learn specific features and predict a group of suitable attributes. Fig. 6 shows our PAL model and Fig. 7 show a sub-network from our PAL model, which is taken from the head region. We also transfer 18-layer Residual Network [32] into our architecture. ResNet18 has 5 layer groups. We apply the conv_0, conv_1 and conv_2 to the first part, conv_3 to the second one and conv_4 for the last one, which is also shown in Fig. 6 and Fig. 7.

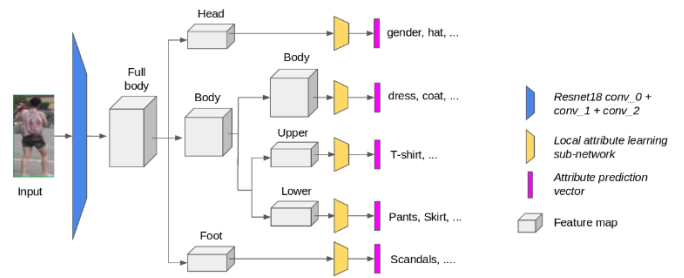

Fig. 6. Our Pedestrian Attribute Learning Model.

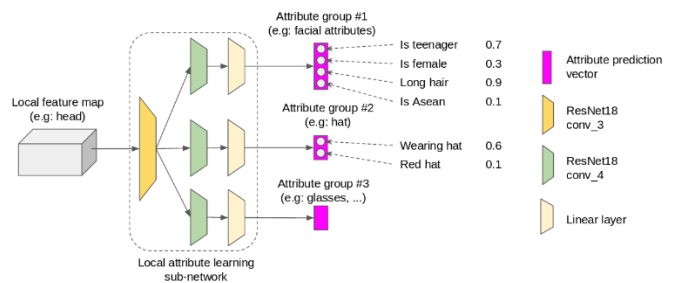

Fig. 7. A Sub-Network from our Pedestrian Attribute Multi-Task Learning Model, which is Taken from the Head Region.

Outputs of the model are a vector whose number of dimensions is equal to number of binary attributes. In the learning stage, we use Binary Cross Entropy function for each attribute, and get average of those of all attributes as our final loss function. With each M-dimension output vector $\hat{a}_i$ and a M-dimension ground truth vector $\hat{a}_i$, where M is the number of binary attributes, then the average Binary Cross Entropy loss function formula is shown in Equation (2):

$$\text{AvgBCELoss} = -1/M \sum_{j=1}^{M} (a_{ij} \log \hat{a}_{ij} + (1 - a_{ij}) \log(1 - \hat{a}_{ij}) \quad (2)$$

*2) Handling imbalanced data:* One of the important improvements of our method is that we incorporate an Imbalance Data Solver into our Person Re-identification system. Imbalanced data is a common problem in classification tasks. There are many ways to handle it, such as: oversampling/ undersampling, use weighted loss function, etc. In the task of multi-label classification like pedestrian attributes recognition, we cannot increase or decrease the amount of samples because it will affect all attributes, and weighting loss will lead to a bunch of hyper-parameters must be tuned. Therefore, we choose the way of adjusting the thresholds of binary attributes, instead of using common values of 0.5. Concretely, for each attribute, we perform a grid search to choose a best threshold from a list of predefined candidate thresholds. The best one is which we get the highest value of Matthews correlation coefficient [43] when using it to convert from probability to binary prediction. Matthews correlation coefficient (MCC) [43] is a famous metric used to measure the quality of a binary classifier. Its formula takes into consideration the 4 popular values of classification problem: true positives (TP), true negatives (TN), false positives (FP) and false negatives (FN), which is shown in Equation (3).

$$\text{MCC=Matthews}_{\text{corrcoef}} = \frac{TPxTN - FPxFN}{\sqrt{(TP+FP)x(TP+FN)x(TN+FP)x(TN+FN)}} \quad (3)$$

Matthews correlation coefficient has value range from −1 to 1:

- It achieves maximum value of 1 when both FP and FN are zeros, which means no sample are false predicted and the classifier result matches exactly the ground truth.

- Contrary, it achieves minimum value of -1 when both of TP and TN are zeros, which means classifier result is completely opposite to the ground truth.

- And it is 0 if the prediction is random.

Therefore, we want to choose which threshold makes the coefficient gain as highest score as possible.

## IV. RESULTS AND EVALUATION

### A. Data Set

We demonstrate our proposed method on a large Re-ID data set named Market1501 [44]. This data set contains 32668 bounding-box images of 1501 persons, which are captured from six different cameras in front of a supermarket near Tsinghua University, China. The authors divided it into three parts: training set has 12936 images of 751 persons and test set has 19281 images of the other 750 persons. From the test set, from 1 to 6 random images of each individual are chosen to form a query set, and remaining are the gallery set. Re-ID task is performed by matching images in query set with images in gallery set, and return top-k most similar ones.

For the pedestrian attribute recognition task, we use this Market1501-attribute data set, which is basically the Market1501 data set, but the attribute annotations are proposed by other authors Lin et al. There are total 27 attributes, but we only use 25 binary attributes which have proportions of positive samples rate more than 0.5%. Table I show these attributes which are clustered into local positions by us.

### B. Using only Person Deep Global Features Learning Model

First, we evaluate our Person Deep Global Features Learning (PDGFL) model when do not use complementary attribute information. We train our PDGFL model in 60 epochs, by Adam optimizer algorithm, with default hyper-parameters, except the learning rate is set to 3.10−4. Input images are rescaled to 192x96 before fed into the network. Feature vector dimensions are 256. We evaluate 3 version of Residual Network [32]: ResNet18, ResNet50 and ResNet101. Results are reported in mAP, top-1, top-5 and top-10 accuracy, which are shown in Table II. In our experiments, ResNet101 achieves highest performance. ResNet50's result is less than ResNet101's by a very small gap, but it has only a half of number of layers compare to ResNet101. This means even more complex and deeper model still cannot distinguish similar appearance individuals.

### C. Attribute Recognition Model

Attribute recognition model is the principle component of our proposed methods. We demonstrate our model in 2 scenarios: without/with Ontology, without/with Matthews correlation coefficient (both are using Ontology). In all attribute recognition experiments, we use the same train-validation-test split and optimizer algorithm as Re-ID experiment, except number of epochs is now set to 10.

Firstly, we re-build the attribute ontology on Market1501-attribute data set. The ontology is shown in Fig. 8.

TABLE I.  25 ATTRIBUTES FROM MARKET1501-ATTRIBUTE DATA SET

| Position | Attribute |
|----------|-----------|
| head | gender, hair length, wearing hat |
| body | carrying backpack, carrying handbag, carrying bag |
| upper | sleeve length, 8 colors of upper clothing |
| lower | length of lower clothing, type of lower clothing, 8 colors of lower clothing |
| foot | none |

TABLE II.  QUERY RESULT OF DIFFERENT MODELS WITHOUT COMPLEMENTARY ATTRIBUTE INFORMATION

| Model | Top-1 | Top-5 | Top-10 | mAP |
|-------|-------|-------|--------|-----|
| ResNet18 | 77.7% | 90.6% | 93.5% | 57.9% |
| ResNet50 | **81.4%** | 91.8% | 94.7% | **65.1%** |
| ResNet101 | **82.0%** | 93.1% | 95.6% | **66.0%** |

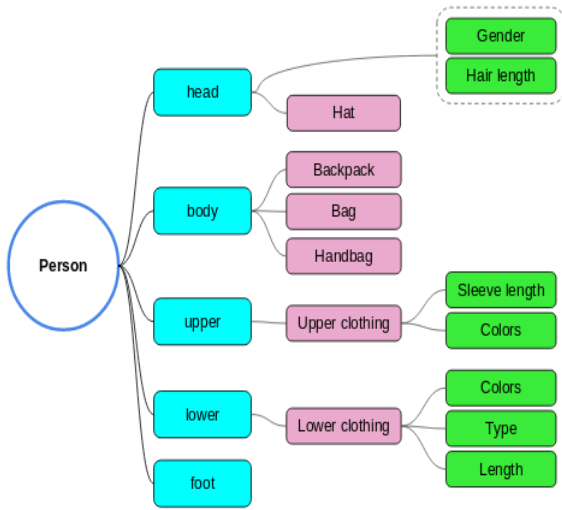

Fig. 8. The PAO Implemented on Market1501-Attribute Data Set.

Secondly, we evaluate our proposed attribute recognition model in four versions:

- Baseline: This is simply a ResNet18 network replaced last 1000-units linear layer by 25-units linear layer. In other words, this model predicts all attributes from a unique global features.

- ONTO: This is the above proposed network, which has multiple branches to predict attributes from suitable local features. But in this version, we still do not handle the imbalance problem.

- ONTO + MCC: This is simply the same as ONTO version, but we handle the imbalance data problem by apply Matthews correlation coefficient in adjusting thresholds. Threshold candidates are predefined by an arithmetic progression from 0.01 to 0.99 with step of 0.01. The best thresholds are then used in test phase to convert continuous outputs to binary values.

- ONTO + MCC + LM: This is the version which has 4 models corresponding to 4 parts: head, body, upper and lower. We observed that, instead of training a unique multi-task model containing all of local branches, training separate models for each local branch make it easy to update new attributes and re-train in the future.

Table III shows the results of 4 versions. We can see some observations:

- With Ontology: As depicted in column Baseline and ONTO, average F1-score increases 12% when using complementary ontology, compare to a plain network. There are 24/25 attribute's which have F1-scores increasing, too. This proofs that using local features, or concretely using ontology, is more powerful than using global features in attribute prediction.

- With Matthews correlation coefficient: As depicted in column ONTO and ONTO + MCC, average F1-score now increases 13% when using MCC. All of attributes have F1-scores increasing, too. Some attributes are improved by a large gap. For example, attribute "wearing hat" has largest growth with 31%, although this attribute positives rate is only 2.6%. This shows that handling imbalance problem by adjusting thresholds help improve significantly the quality of a multi-label classifier.

- With Local multi-task training: As depicted in column ONTO + MCC and ONTO + MCC + LM, average F1-score again increases 13% when training multiple models corresponding to each local part. And all of attributes have F1-scores increasing, too. Moreover, there are many attributes having result greater than 90%, includes: up white, up red, up yellow, lower type, and down black. This shows that, training separate models for each position can also improve the prediction result.

### D. Attribute Filtering for Person Re-Identification

*1) Compare to the case of using only global deep features:* From results of attribute recognition models in Table III, because of not all of attributes give remarkable scores, instead of use all of them, we only choose five highest F1-score attributes to improve re-id performance. Concretely, for each query image in query set of Market1501 data set:

*a) Firstly:* We use each of these attributes to filter out candidates in gallery set that mismatch the attributes of query image.

TABLE III. RESULTS OF DIFFERENT VERSIONS OF PROPOSED MODEL (F1-SCORES)

| Position | Attribute | Baseline | ONTO | ONTO + MCC | ONTO + MCC + LM |
|---|---|---|---|---|---|
| Head | gender | 40.26 | 71.45 | 78.03 | 88.91 |
| | hair length | 51.32 | 65.62 | 73.43 | 86.38 |
| | wearing hat | 06.74 | 22.82 | 53.42 | 68.26 |
| Body | backpack | 45.89 | 53.06 | 70.12 | 84.76 |
| | bag | 43.57 | 46.94 | 53.54 | 77.10 |
| | handbag | 11.40 | 32.47 | 54.57 | 64.77 |
| Upper | sleeve length | 11.96 | 44.86 | 66.57 | 74.94 |
| | up black | 47.63 | 69.53 | 83.06 | 88.75 |
| | up white | 73.58 | 74.05 | 79.58 | 91.92 |
| | up red | 71.47 | 73.88 | 88.93 | 90.05 |
| | up purple | 20.02 | 47.74 | 67.91 | 84.99 |
| | up yellow | 72.86 | 76.80 | 81.37 | 94.16 |
| | up gray | 21.63 | 52.53 | 68.89 | 82.32 |
| | up blue | 45.84 | 58.23 | 68.91 | 84.92 |
| | up green | 41.53 | 44.68 | 64.22 | 84.99 |
| Lower | lower length | 59.56 | 74.41 | 80.85 | 83.26 |
| | lower type | 78.78 | 82.20 | 89.20 | 92.16 |
| | down black | 74.19 | 83.67 | 87.85 | 91.19 |
| | down white | 48.63 | 49.35 | 69.58 | 78.10 |
| | down pink | 62.77 | 61.80 | 66.79 | 89.07 |
| | down yellow | 0.21 | 11.08 | 31.34 | 43.08 |
| | down gray | 51.20 | 57.92 | 64.61 | 78.84 |
| | down blue | 58.77 | 61.43 | 64.64 | 81.34 |
| | down green | 29.29 | 44.52 | 54.96 | 76.30 |
| | down brown | 43.85 | 53.61 | 68.46 | 82.39 |
| | **Average** | **43.32** | **56.59** | **69.20** | **81.72** |

*b) Secondly:* the remaining candidates are used to find K nearest neighbors by comparing Euclidean dissimilarity distance of deep features extracted by the above ResNet50 Re-ID model.

The 5 chosen attributes are: up red, up white, up yellow, lower length and down black. Table IV shows attributes pre-filtering results in mAP and top-K. As depicted in Table IV, pre-filtering by attribute down black gives best results in top-K accuracy, and by attribute up red gives best result in mAP. Consider all of these 5 attributes in pre-filtering, although the top-k accuracy values are improved by a small gap, the mAP values increase remarkably, at least 9.3% comparing to the case that does not use attributes in pre-filtering. This indicates that, if in the future we can have better attribute recognition models, so that more attributes have high prediction results, then Re-ID results would be derived to a better performance too.

*2) Compare to the case of using global and local deep features:* Most of previous works use attributes as auxiliary task for enhancing the global/local deep features, and do not employ attributes prediction in the test step. Therefore, we perform some experiments to compare the two cases: using global and local deep features with using global deep features and attributes information. Combination of global and local deep features in our experiments are extracted as follow:

*a) For each position:* we get the output of the local part in our above network (Fig. 6), which is a feature map.

*b) Then:* we apply a max-pooling operation to convert feature map to a feature vector. This is the local deep feature vector of the corresponding position.

*c) Finally:* we concatenate deep global features and local feature of one of the 4 regions (head, body, upper, lower) to form a unique deep feature vector and then use it to perform matching process in test stage.

Combination of deep global features and attributes information is exactly the pre-filtering strategy in the previous section. For each position, we compare two cases: i) use all attributes of that position; and ii) use only one attribute with best prediction of that position. Results of the comparisons of all of 4 positions are showed in Table V, the arrows indicates the result of using attribute information is higher or lower than using local features, and the bold values is the highest values between 3 cases in that positions.

As depicted in Table V, when using all attributes in filtering step, top-k accuracies and mAP in all positions are lower than combinations of global and local features. However, it is the opposite for the case of using only one best attribute. In the position head, attribute gender has higher the top-1, top-5 and mAP at about 3-5%, and the top-10 accuracy is smaller only 0.1% compare to using complementary local features. In the position body, attribute backpack is not as good as local features, because it has a not-too-good prediction F1-score, about 84%. In the positions upper and lower, using corresponding best attributes gives performance totally higher than deep local features. Fig. 9 shows some samples that query results are rearranged and improved by attribute filtering.

TABLE. IV.    QUERY RESULTS BY ATTRIBUTES PRE-FILTERING USING 5 ATTRIBUTES HAVING BEST PREDICTION

| Attribute Pre-filtering | Top-1 | Top-5 | Top-10 | mAP |
|---|---|---|---|---|
| None | 81.4% | 91.8% | 94.7% | 65.1% |
| up red | ↑83.3% | ↑93.9% | ↑92.6% | **↑75.2%** |
| up white | ↑83.6% | ↑93.7% | ↑95.8% | ↑74.4% |
| up yellow | ↑83.8% | ↑93.6% | ↑95.9% | ↑74.9% |
| lower type | ↑84.2% | ↑94.1% | ↑96.0% | ↑74.7% |
| down black | **↑85.2%** | **↑95.3%** | **↑96.9%** | ↑74.8% |

TABLE. V.    COMPARISON BETWEEN COMBINATION OF GLOBAL AND LOCAL DEEP FEATURES AND COMBINATION OF GLOBAL DEEP FEATURES AND ATTRIBUTE INFORMATION

| Case | Top-1 | Top-5 | Top-10 | mAP |
|---|---|---|---|---|
| *Position: Head* | | | | |
| Deep Global + Local Feature | 78.4% | 89.7% | 93.9% | 56.8% |
| Deep Global feature + All local attributes | ↓75.1% | ↓87.2% | ↓92.6% | ↓53.3% |
| Deep Global feature + Only attribute *gender* | **↑81.6%** | **↑91.1%** | ↓93.8% | **↑61.2%** |
| *Position: Body* | | | | |
| Deep Global + Local Feature | **80.4%** | **89.7%** | 94.9% | 62.0% |
| Deep Global feature + All local attributes | ↓72.7% | ↓79.6% | ↓85.1% | ↓50.3% |
| Deep Global feature + Only attribute *backpack* | ↓78.3% | ↑91.4% | ↑95.5% | **↑63.1%** |
| *Position: Upper* | | | | |
| Deep Global + Local Feature | 80.3% | 91.5% | 94.8% | 61.7% |
| Deep Global feature + All local attributes | ↓77.7% | ↓89.4% | ↓92.9% | ↓59.9% |
| Deep Global feature + Only *up yellow* | **↑83.8%** | **↑93.6%** | **↑95.6%** | **↑74.9%** |
| *Position: Lower* | | | | |
| Deep Global + Local Feature | 78.3% | 90.9% | 94.5% | 61.1% |
| Deep Global feature + All local attributes | ↓69.1% | ↓77.4% | ↓82.5% | ↓47.8% |
| Deep Global feature + Only attribute *lower type* | ↑84.2% | ↑94.1% | ↑96.0% | ↑74.7% |

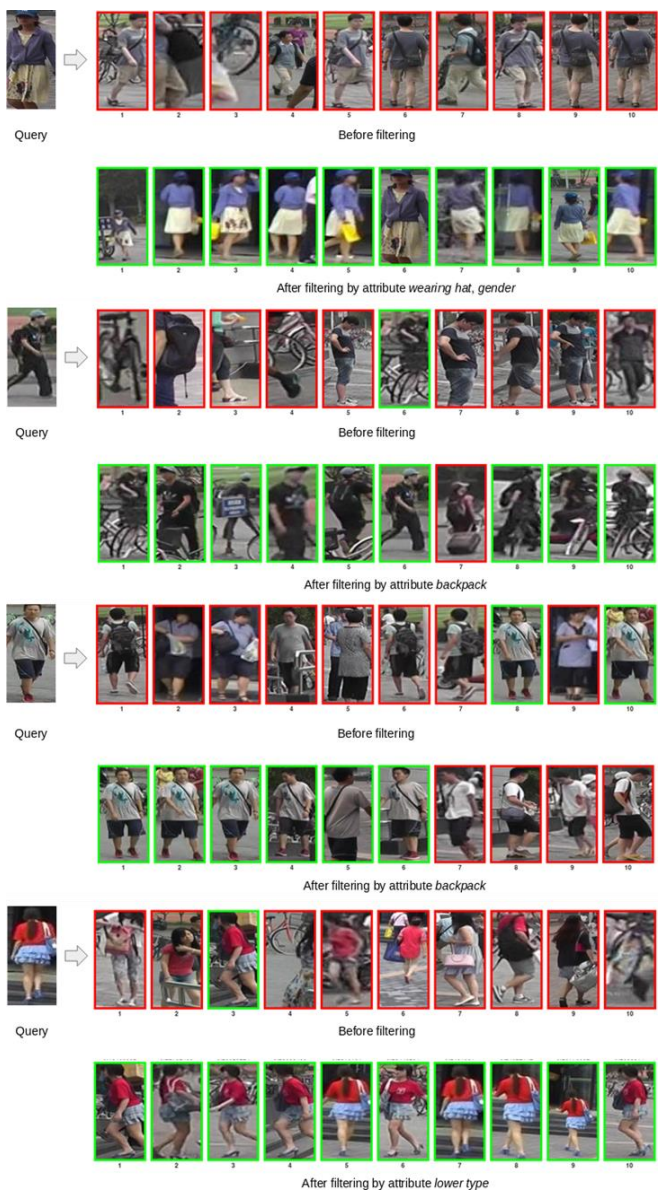

Fig. 9. Some Samples that Query Results are Improved by Attribute Filtering.

*3) Compare to other methods:* We use attribute "down black" which has the best performance in improving Re-ID results in comparison with related works. As depicted in Table VI, our method achieves the higher performance than the other works in mAP, top-5 and top-10 accuracy. Note that these works only use attribute as auxiliary information in learning stage. The results show that using attribute as a pre-filter in inference stage can achieve equivalent or even better performance. The methods presented in [24], [26], [28] were selected to compare performance with our method for the following reasons: all these methods have used pedestrian's attributes in Person Re-Identification in different ways but have not yet used attribute pre-filters and considered data imbalance. We would like to show that our method can overcome their drawbacks listed in Section II (part C).

TABLE VI. COMPARISON WITH OTHER METHODS ON MARKET1501 DATA SET

| Methods | Top-1 | Top-5 | Top-10 | mAP |
|---|---|---|---|---|
| Schumann and Stiefelhagen [24] | 83.61% | 92.61% | 95.34% | 62.6% |
| Lin et al. [26] | 84.29% | 93.2% | 95.19% | 64.67% |
| Zhang and Xu [28] | **86.58%** | 94.48% | 96.73% | 68.08% |
| Ours, pre-filtering by attribute *down black* | 85.2% | **95.3%** | **96.9%** | **74.8%** |

## V. CONCLUSIONS

In this paper, we present a new method using semantic information like as pedestrian attributes to improve person re-identification performance. Our methods is a unified Re-ID system consisted of two main modules: 1) Pedestrian Attributes Learning model (PAO + Local MDCNN + IDS); 2) Person Re-ID model (Deep Global Features based Person Re-ID + Pedestrian Attribute based Person Re-ID). We show that the performance of our Re-ID system is better than some state-of-the-art Re-ID methods. In the future, if more powerful attribute recognition model were proposed, Re-ID task would be driven to a better performance and Re-ID system at semantic level will be integrated to Visual Question Answering (VQA) to improve the intelligence of video surveillance system.

## ACKNOWLEDGMENT

This research is funded by Viet Nam National University Ho Chi Minh City (VNUHCM) under grant no. B2018-18-01. Thank to AIOZ Pte Ltd. company for the valuable support on internship cooperation.